\pdfoutput=1
\documentclass[11pt]{article}
\usepackage[symbol]{footmisc}
\usepackage{authblk} % 用于作者和单位管理
\usepackage[utf8]{inputenc} % 支持特殊字符
\usepackage{multirow}
\usepackage{subcaption}
\usepackage[table,dvipsnames]{xcolor}
\usepackage[]{EMNLP2023}
\usepackage{graphicx}
\usepackage{times}
\usepackage{latexsym}
\usepackage{amsmath}
\usepackage{amssymb}

\usepackage{geometry}
\usepackage{float}
\usepackage{newtxtext}
\usepackage{tcolorbox}
\usepackage[T1]{fontenc}
\usepackage[utf8]{inputenc}
\usepackage{booktabs}
\usepackage{microtype}

\usepackage{inconsolata}

\title{Fast on the Easy, Deep on the Hard: Efficient Reasoning via Powered Length Penalty}

\author[1,3*]{\textbf{Zehui Ling}}
\author[1,2,3*]{\textbf{Deshu Chen}}
\author[1,2,3]{\textbf{Hongwei Zhang}}
\author[3]{\textbf{Yifeng Jiao}}
\author[3]{\textbf{Xin Guo}}
\author[1,3 $\dagger$]{\textbf{Yuan Cheng}}
\affil[1]{Artificial Intelligence Innovation and Incubation Institute, Fudan University}
\affil[2]{School of Data Science, Fudan University}
\affil[3]{Shanghai Academy of Artificial Intelligence for Science}
\affil[$\dagger$]{Corresponding author:{cheng\_yuan@fudan.edu.cn}}
\begin{document}
\maketitle
% \begin{abstract}
% The reasoning capabilities of large language models (LLMs) have shown remarkable progress, achieving strong performance across a variety of challenging benchmarks. To further enhance reasoning, techniques such as Chain-of-Thought prompting have been proposed. However, these methods often lead to substantially longer outputs, resulting in increased computational latency. While some approaches attempt to reduce reasoning length via reinforcement learning, they typically apply uniform penalties regardless of the problem's complexity. In this work, we aim to improve the efficiency of LLM reasoning by encouraging brevity on simple problems and preserving accuracy on complex ones, thereby enhancing the model’s overall usability. Specifically, we control the efficiency of reasoning of the model by segmenting the reward function and incorporating a penalty term of length. Our method has achieved remarkable results in benchmark tests on three datasets: GSM8K, Math500, and AIME2024. In the relatively simple datasets GSM8K and Math500, our approach has successfully reduced the output length while maintaining or even improving accuracy. In the more challenging AIME2024 dataset, our method has led to an increase in accuracy.
% \end{abstract}

\begin{abstract}
Large language models (LLMs) have demonstrated significant advancements in reasoning capabilities, performing well on various challenging benchmarks. Techniques like Chain-of-Thought prompting have been introduced to further improve reasoning. However, these approaches frequently generate longer outputs, which in turn increase computational latency. Although some methods use reinforcement learning to shorten reasoning, they often apply uniform penalties without considering the problem's complexity, leading to suboptimal outcomes. In this study, we seek to enhance the efficiency of LLM reasoning by promoting conciseness for simpler problems while preserving sufficient reasoning for more complex ones for accuracy, thus improving the model's overall performance. Specifically, we manage the model's reasoning efficiency by dividing the reward function and including a novel penalty for output length. Our approach has yielded impressive outcomes in benchmark evaluations across three datasets: GSM8K, MATH500, and AIME2024. For the comparatively simpler datasets GSM8K and MATH500, our method has effectively shortened output lengths while preserving or enhancing accuracy. On the more demanding AIME2024 dataset, our approach has resulted in improved accuracy.

\end{abstract}
\footnotetext{*Equal Contribution}
\footnotetext{$\dagger$The corresponding author}
\section{Introduction}
Large language models have made astonishing progress recently, especially in their reasoning capabilities \cite{zhou2022least,chen2024language,suzgun2022challenging}. However, the enhancement of reasoning capabilities comes at the cost of a significant increase in additional overhead. For example, Chain-of-Thought (CoT) \cite{nye2021show,wei2022chain} typically achieved by adding "Let's think step by step" to the input prompt, breaks down the steps of problem-solving through a chain of reasoning, ultimately arriving at a correct answer. CoT is highly effective, as a result, many existing models, such as DeepSeek-R1 \cite{guo2025deepseek}, uses CoT data to perform reinforcement learning on the models, achieving amazing results.
\par
\begin{tcolorbox}
    [colback = Emerald!10, colframe = cyan!40!black, title = Question,fontupper=\small]
     Keegan was running a car wash with his friend Tashay to raise money for a baseball camp. They needed to raise \$200 for the two of them. By 3 pm, Keegan had earned \$83 and Tasha had earned \$91. How much more did they need to earn to reach their goal?
\end{tcolorbox}

\begin{tcolorbox}
    [colback = Emerald!10, colframe = cyan!40!black, title =Original Model (265 tokens),fontupper=\small]
     First, I need to determine the total amount Keegan and Tashay have earned so far by adding their individual earnings. Keegan earned \$83, and Tashay earned \$91. Adding these together gives: 83 + 91 = 174 Next, I'll subtract their total earnings from the goal of \$200 to find out how much more they need to earn: 200 - 174 = 26 Therefore, they still need to earn \$26 to reach their goal. \underline{</think>} To determine how much more Keegan and Tashay need to earn to reach their goal of \$200, follow these steps: 1. **Calculate the total amount earned so far:** Keegan's earnings + Tashay's earnings=
     \textbf{\$83 + \$91 = \$174} 2. **Determine the remaining amount needed to reach the goal:** \[ \textbf{Goal} - \textbf{Total earned} = \textbf{\$200 - \$174 = \$26} \] Therefore, they need to earn an additional \(\boxed{\$26}\) to reach their goal.
\end{tcolorbox}

\begin{tcolorbox}
    [colback = Emerald!10, colframe = cyan!40!black, title = Ours (54 tokens),fontupper=\small]
     First, add Keegan's and Tashay's earnings: \textbf{\$83 + \$91 = \$174}. Subtract this from their goal: \textbf{\$200 - \$174 = \$26}. \boxed{26}
\end{tcolorbox}

\begin{figure}[htbp] % 使用 figure* 环境跨越两栏
    \begin{subfigure}[b]{0.5\textwidth}
    \end{subfigure}
    \caption{Comparison of model outputs under the same prompt. Our method produces a shorter yet accurate response, demonstrating more efficient reasoning.}
    \label{fig:1}
\end{figure}

Although the method of generating CoT is highly effective in solving complex and difficult problems, this approach also incurs significant overhead when applied to solving simple problems \cite{feng2023alphazero}. For example, when solving a simple problem like "3 + 4 = ?", Deepseek-R1, with its deep thinking capability, would first spend several seconds pondering and consume a large number of tokens before arriving at the answer. Such significant delay and resource consumption is something we would like to avoid.
Recently, several approaches have been proposed with the goal of shortening the reasoning path required by models when answering simple questions \cite{sui2025stop}. A wide range of methods have been proposed to achieve this objective. 

(1) \textbf{Prompt-guided Efficient Reasoning} \cite{han2024token,xu2025chain,ding2024break,lee2025well} focuses on designing specialized prompts to guide models in generating shorter, more direct reasoning paths to answer questions efficiently. One advantage is that it does not require additional model training. However, prompt-based control tends to be less effective for models with smaller parameter sizes, where the controllability and performance may fall short of expectations. 

(2) \textbf{Variable-Length CoT} \cite{xia2025tokenskip,ye2025limo,ma2025cot,munkhbat2502self,liu2024can} trains large language models using supervised fine-tuning on datasets that contain reasoning chains of varying lengths, allowing the model to adapt its reasoning depth based on the complexity of the input. One advantage lies in its ability to fine-tune models by constructing tailored datasets, which demands fewer computational resources and achieves strong performance within the training data domain. However, it exhibits limited generalization ability and only moderate performance on examples outside the training distribution. 

(3) \textbf{Length Reward Designing} \cite{team2025kimi,arora2025training,luo2025o1,qu2025optimizing} shapes the reward function in reinforcement learning to encourage models to generate shorter reasoning paths by assigning higher rewards to concise and correct answers while penalizing overly long or incorrect responses. It requires greater computational resources; however, it endows the model with stronger generalization capabilities after training and does not rely on additional prompts or markers during inference.

% The advantage of using the Prompt-guided Efficient Reasoning method is that it does not require additional training of the model. However, the prompt control method is difficult to manage in models with smaller parameter sizes, and the effect is not ideal. 

% The advantage of using the SFT with Variable-Length CoT Data method lies in fine-tuning through creating datasets, which requires less computational resources and performs well in the training data domain. However, it lacks generalization ability and shows mediocre performance on tests outside the training data domain. 

% RL with Length Reward Design requires more computational resources, but this method enables the model to have stronger generalization ability after training, and no additional markers are needed when asking questions. 

Although these methods all aim to reduce the length of reasoning chains in language models, they generally overlook variations in the difficulty of the question. Consequently, such approaches may inadvertently compromise the model's ability to handle complex or challenging tasks. This raises an important research question:

\textit{Can we teach language models to think like humans—fast on the easy, deep on the hard?}

Intuitively, difficult problems often require more reasoning and problem-solving steps \cite{kahneman2011thinking,perez-etal-2020-unsupervised}. To arrive at correct answers, large language models typically expend more tokens on challenging questions. Motivated by this observation, we propose using model response length as an indicator of question difficulty. Our goal is to develop a method that enables models to reduce resource consumption on simple questions while maintaining high accuracy on complex ones.

We achieve this by adapting the RLOO algorithm \cite{ahmadian2024back} with a modified reward function. Specifically, we introduce a length penalty into the reward structure, such that the model receives the highest reward for short and correct responses, a slightly lower reward for longer but still correct responses, and no reward for incorrect answers. This encourages the model to generate concise and accurate reasoning paths, adapting its computation based on question difficulty,as shown in Figure \ref{fig:1}
We observe that even when the model is trained exclusively on the relatively simple GSM8K dataset, it exhibits strong generalization to more challenging math benchmarks.
We evaluate our approach on two model variants: DeepSeek-R1-Distill-Qwen-1.5B and DeepSeek-R1-Distill-Qwen-7B. For the 1.5B model, our method achieves notable improvements. On the GSM8K dataset, it reduces output token count by 40\% while increasing accuracy by 10\%. On the more challenging AIME2024 dataset, the model reduces token usage by 15\%, with a corresponding 4.8\% gain in accuracy. In the case of the 7B model, our method demonstrates even greater efficiency. It reduces token usage on GSM8K by as much as 90\%, with only a minor accuracy drop of 1.4\%. On AIME2024, when token count is reduced by 20\%, the accuracy remains virtually unchanged.

In conclusion, our contributions can be summarized as follows:
\begin{itemize}
\item[$\bullet$] We propose a method based on a novel reward function that effectively imposes length penalties on simple questions while imposing almost no length penalties on difficult ones.
\item[$\bullet$] By analyzing the advantage function, we demonstrate that our approach better aligns with the principle of "fast on the easy, deep on the hard" compared to existing reinforcement learning methods for compressing Chain-of-Thought reasoning.
\item[$\bullet$] Extensive experiments validate the effectiveness of our method, and in-depth analyses provide valuable insights for future research in this area.
\end{itemize}

\section{Related Work}
\begin{figure*}[htbp]\label{Figure}
\centering
    \begin{subfigure}[b]{0.49\textwidth} % 调整为0.48留出间距
        \includegraphics[width=\linewidth]{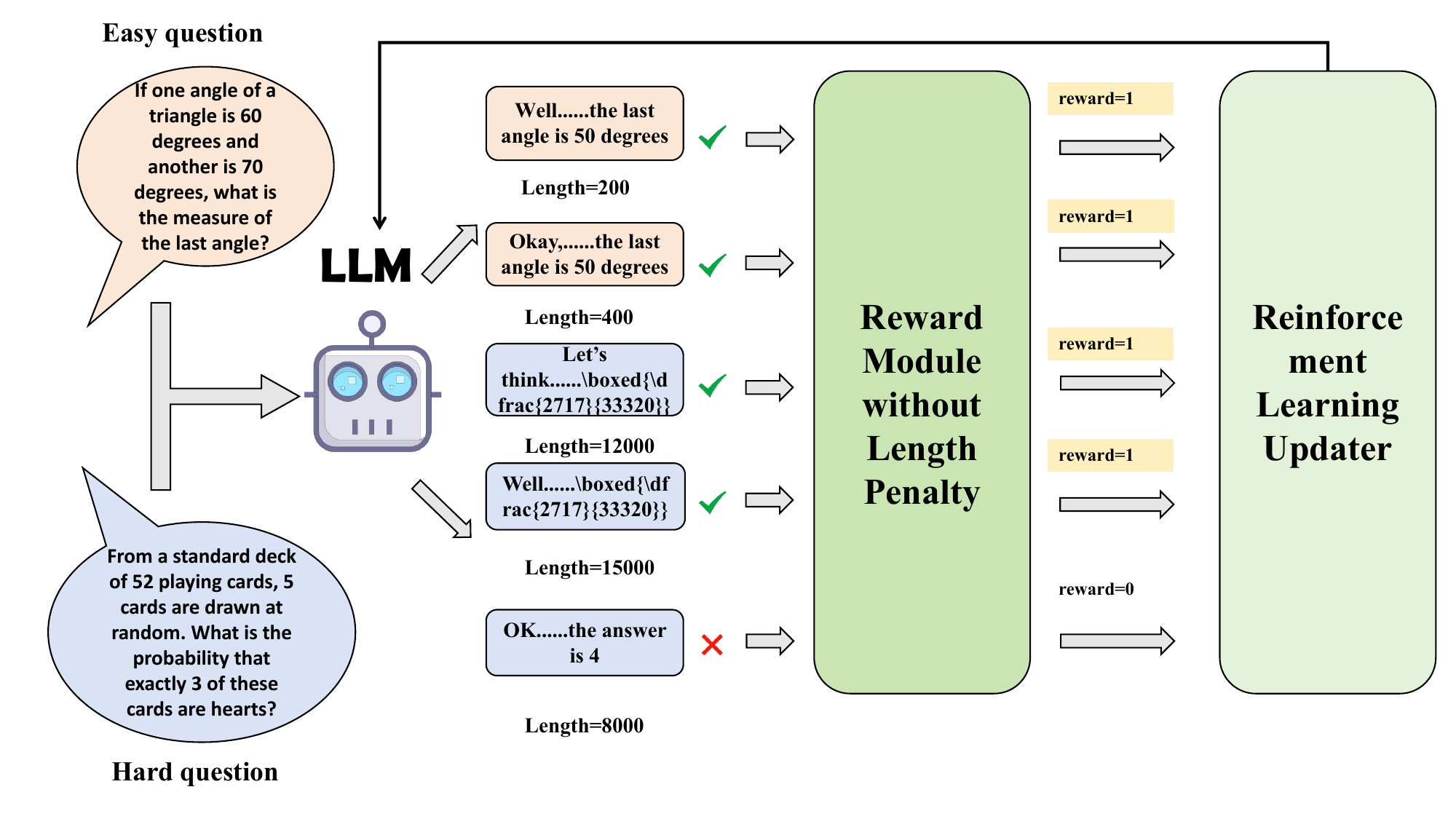}
    \caption{RL without length penalty.}
    \end{subfigure}
    \hfill % 添加水平填充，确保两个图片之间有间隔
    \begin{subfigure}[b]{0.49\textwidth} % 调整为0.48留出间距
        \includegraphics[width=\linewidth]{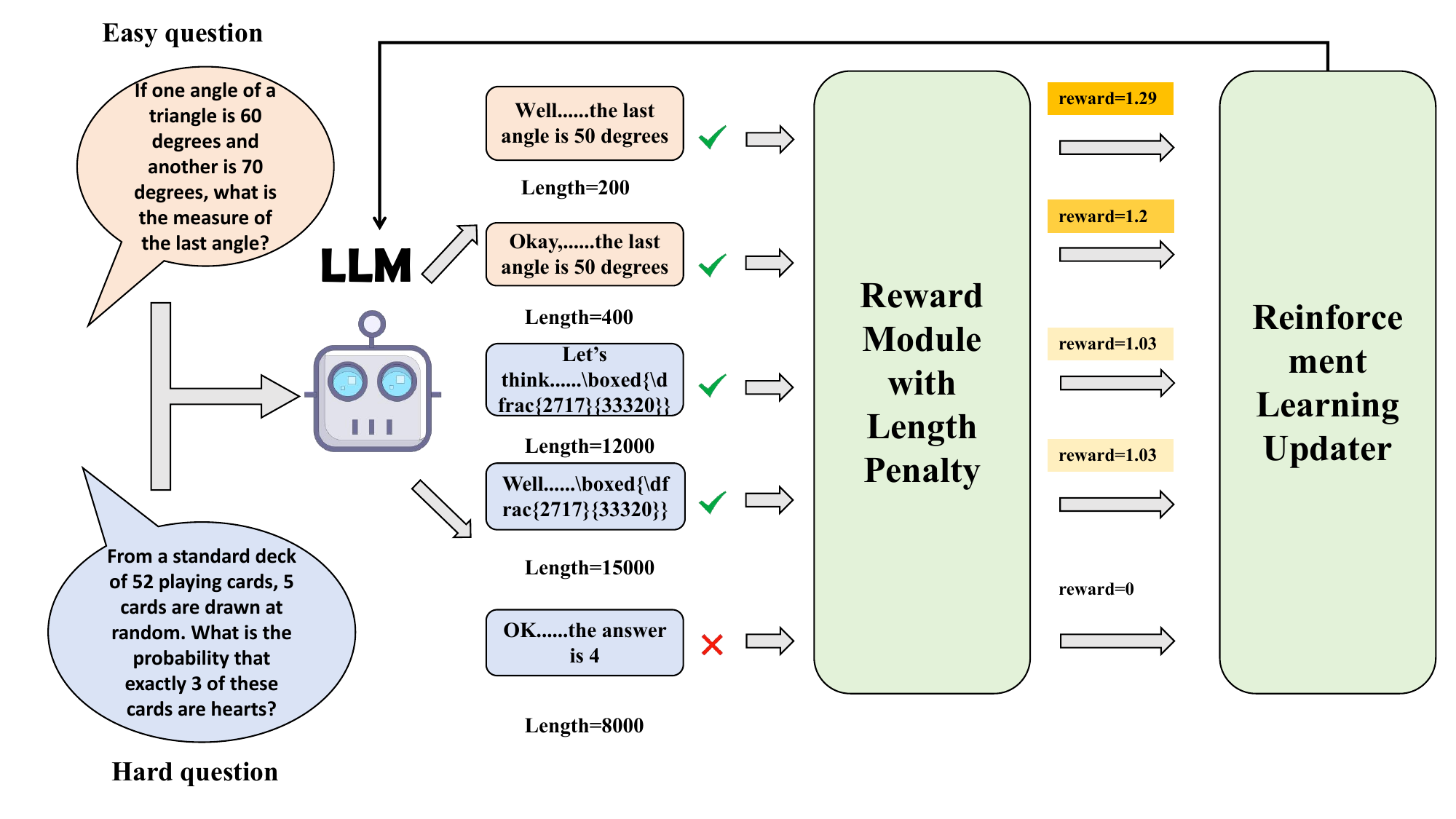}
        \caption{RL with length penalty($\alpha=4,\gamma=0.5$).}
        \label{fig:sub3}
    \end{subfigure}
    \caption{Difference between original RL and ours. The same color represents the question corresponding to the answer.In our method,
the model adapts the length penalty according to the difficulty of the problem: a high penalty is imposed for simple tasks to encourage concise responses, whereas the penalty is minimized for complex tasks to permit more comprehensive answers. Incorrect outputs receive 0 reward irrespective of response length.}
    \label{fig:pipeline}
\end{figure*}
\textbf{Large Reasoning Model} In recent years, large language models have demonstrated remarkable potential in tackling complex reasoning tasks. Pioneering approaches like Chain-of-Thought and Tree-of-Thought \cite{yao2023tree} have enabled these models to perform multi-step logical inference through step-by-step reasoning processes. Meanwhile, models such as OpenAI \cite{achiam2023gpt}, Deepseek-R1 \cite{guo2025deepseek}, QwQ-preview \cite{qwq-32b-preview}, and Kimi \cite{team2025kimi} have further enhanced their reasoning abilities via large-scale reinforcement learning, equipping them with advanced skills like branching and validation, which substantially boost their overall reasoning performance.\\
\textbf{Efficient reasoning} Although reasoning models and the Chain-of-Thought approach enhance model reasoning capabilities, they also entail increased computational costs. To address this, various methods have been proposed to improve reasoning efficiency. Token-Budget \cite{han2024token} estimates the minimum number of tokens needed to answer a question and incorporates a prompt to guide the model to keep its reasoning within this token limit, thereby reducing computational overhead. However, this method incurs additional token usage when estimating the problem’s difficulty. TokenSkip \cite{xia2025tokenskip} evaluates the importance of each token in the response, marks them accordingly, and fine-tunes the model on a dataset where a portion of less important tokens are removed. This often leads to answers that omit conjunctions, resulting in less coherent responses. Another approach, training language models for efficient reasoning \cite{arora2025training}, modifies the reward function in RLOO by adding a length penalty based on the number of actions $\alpha$, followed by reinforcement learning on a mixed dataset. However, it does not differentiate the length penalty between difficult and easy samples.

% Existing methods for enhancing reasoning efficiency can be broadly categorized into three main approaches: Prompt-Guided Efficient Reasoning, SFT with Variable-Length CoT, aznd RL Optimization via Length Reward. 

\section{Methodology}

\subsection{Overview}
Our goal is to enable the model to perform efficient reasoning, that is, to produce accurate answers while minimizing token usage. For simple questions, the model is encouraged to generate concise responses. For more challenging questions, it is allowed to prioritize accuracy over brevity. To achieve this, we apply reinforcement learning with a carefully designed reward function that guides the model toward this balance of efficiency and correctness.Our pipeline can be seen in Figure \ref{fig:pipeline}
\subsection{Problem Setup}
We consider a large language model parameterized by $\theta$, denoted as $\pi_\theta$. Given an input sequence $x$, the model generates a response $y$, where $y$ is sampled from the conditional distribution $\pi_\theta(\cdot \mid x)$. Since our evaluation involves problems with objective, verifiable answers, we define the accuracy as follows:
\begin{equation}\label{accuracy_eq}
\text{accuracy} = 
\begin{cases} 
1, & \text{if $y=y^{\ast}$,} \\
0, & \text{otherwise.}
\end{cases}
\end{equation}
where $y^*$ denotes the correct answer, and $y$ is the model’s generated answer. Only the final answer needs to match the ground truth; the intermediate reasoning steps or the complete output sequence need not be identical.

\subsection{RL with Length Cliping}
In reinforcement learning, the Proximal Policy Optimization (PPO) algorithm \cite{schulman2017proximal} is one of the most widely used approaches. However, PPO introduces substantial memory overhead due to its reliance on both a critic model and an external reward model. To avoid this limitation, we adopt the simpler REINFORCE framework, which eliminates the need for a critic model and thereby significantly reduces memory consumption. Furthermore, since our task involves mathematical problems with deterministic answers, the accuracy function defined in Eq. (\ref{accuracy_eq}) can naturally serve as the reward signal. As a result, we also omit the reward model.

The gradient update in the Reinforce framework is given by:
\begin{equation}
\mathbb{E}_{x \sim \mathcal{D}, y \sim \pi_\theta(.|x)} \left[ R(y, x) \nabla_\theta \log \pi_\theta(y|x) \right],
\end{equation}

In order to reduce the bias, we employ the RLOO method. We generate $k$ samples for each answer and calculate the advantage value by subtracting the average return of the remaining $k - 1$ samples.

\begin{small} 
\begin{equation}
\frac{1}{k} \sum_{i=1}^{k} \left[ R(y^{(i)}, x) - \frac{1}{k-1} \sum_{j \neq i} R(y^{(j)}, x) \right] \nabla \log \pi(y^{(i)} | x),
\end{equation}
\end{small}
$y^{(i)}$ is the independent sample generated from $\pi_{\theta}(\cdot \mid x)$.
To ensure stable training, we set a maximum token limit for generation and include the prompt: \textit{"Please reason step by step, and put your final answer within boxed{}."} This setup ensures that if the output is too long, the final answer will not be produced, which is then considered incorrect. We assign a reward for correct answers, while incorrect answers receive a reward of zero. Our reward function is defined as follows:
\begin{equation}
\begin{small}
R(y^{(i)}, x)=
\begin{cases} 
f(\text{len}(y^{(i)}), & \text{if $y=y^{\ast}$,} \\
0, & \text{otherwise}.
\end{cases}
\end{small}
\end{equation}
\subsection{Length Penalty}
Since binary correctness alone does not capture the efficiency of reasoning, we introduce a length-based penalty to encourage concise solutions. Given the answer $y^{(i)}$, denote $\text{len}({y}^{(i)})$ as the length of the $i$-th predicted sequence. The Powered Length Penalty (PLP) formula we define is as follows:
\begin{equation}
f(\mathrm{len}(y^{(i)})) = 1 + \frac{\alpha}{{\text{len}(y^{(i)})}^\gamma},
\end{equation}
where $\alpha\geq0, \gamma>0$ is the hyperparameter. Under PLP, longer responses receive relatively smaller penalties, while shorter responses are penalized more heavily. This design aligns with our desired behavior: favoring brevity for simple questions and prioritizing accuracy for complex ones. We choose PLP over a standardized penalty approach because the latter tends to normalize response lengths across different question difficulties, thereby reducing sensitivity to the actual length of the generated answers. As a result, it becomes difficult to control the trade-off between reasoning efficiency and correctness. In contrast, PLP allows for finer-grained control by directly incorporating response length into the penalty term, enabling the model to adapt its reasoning depth based on question complexity.
\begin{figure*}[htbp]\label{Figure}
\centering
    \begin{subfigure}[b]{0.48\textwidth} % 调整为0.48留出间距
        \includegraphics[width=\linewidth]{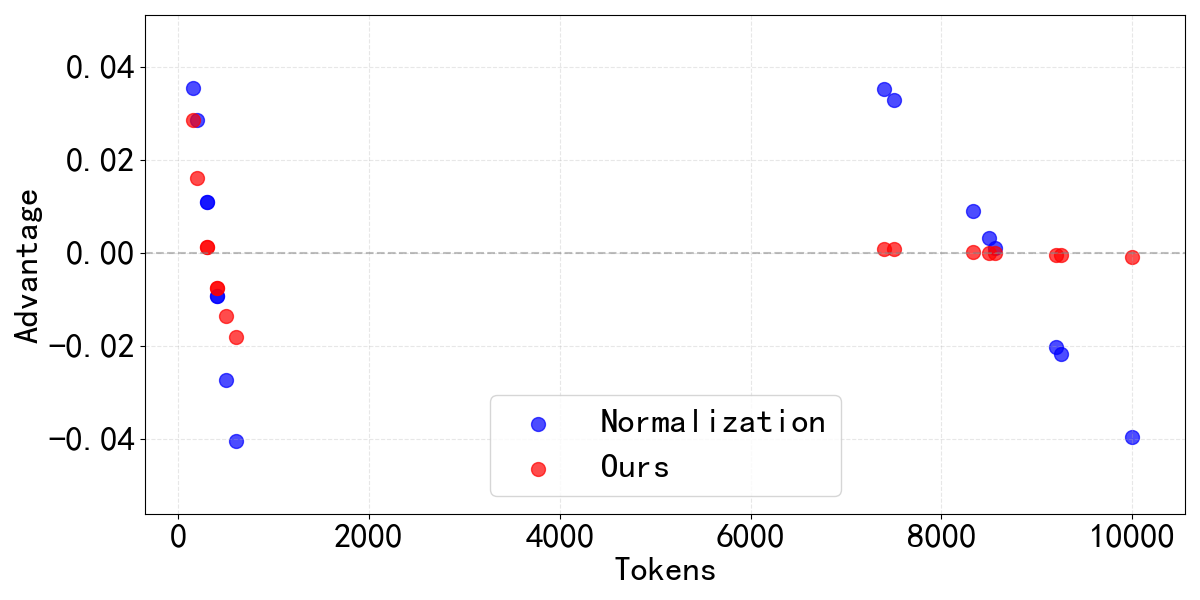}
        \caption{Advantage when all responses are correct.}
        \label{fig:sub1}
    \end{subfigure}
    \hfill % 添加水平填充，确保两个图片之间有间隔
    \begin{subfigure}[b]{0.48\textwidth} % 调整为0.48留出间距
        \includegraphics[width=\linewidth]{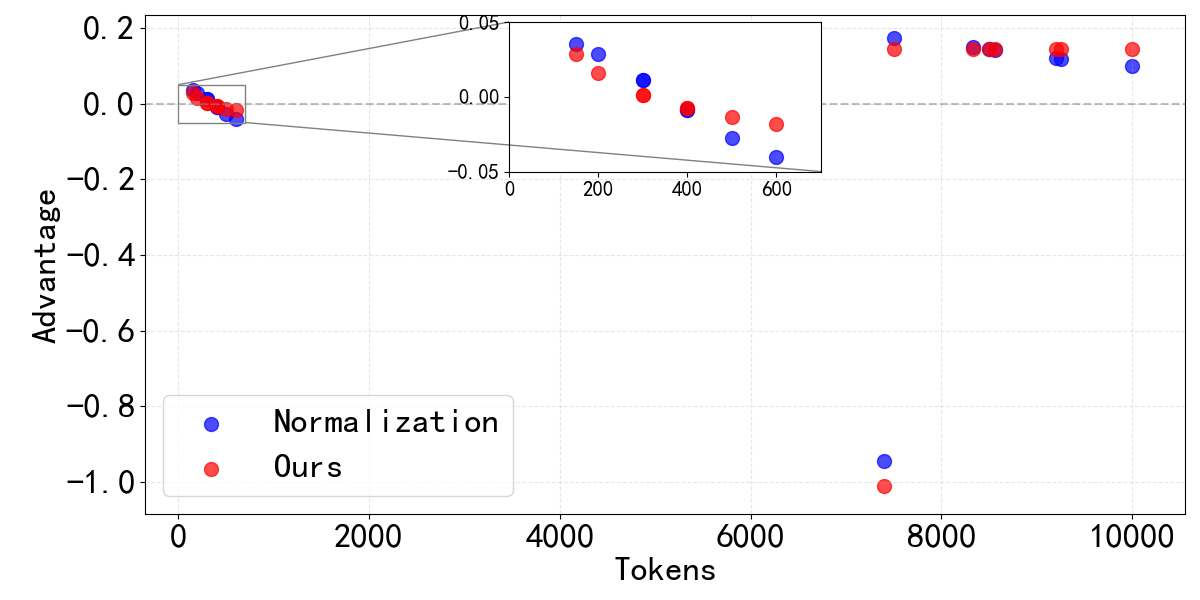}
        \caption{Advantage when one response is wrong.}
        \label{fig:sub3}
    \end{subfigure}
    \caption{Comparison between standardized and absolute length penalty methods across two example ranges: 300–600 and 7,000–10,000 tokens. Blue indicates the standardized method, while red denotes the absolute method.}
    \label{fig:main}
\end{figure*}
% Through ILP, longer responses receive smaller penalties, while shorter responses are penalized more heavily. It fits the pattern of 简单问题追求简洁答案, 复杂问题相较之追求准确率. Thus ILP is chosen rather the standardized penalty since the standardized penalty scheme is not sensitive to the length of the answer. 

% In this function, when dealing with simple questions, the accuracy of the answers is relatively high. In such cases, we expect the length of the answers to be shorter, so the penalty for the length will be greater. When encountering difficult problems, the steps required to solve the problems become more numerous, so the length of the answers is also longer. At this time, in order to pursue accuracy, we need to reduce the length penalty. We adopt the absolute length penalty instead of the standardized penalty scheme, because when calculating the reward, the standardized penalty scheme is not sensitive to the length of the answer. We will demonstrate in the analysis of the advantage function in the next section that this approach is more effective than the method of penalizing with standardized length.

\section{Understanding the Advantage Function}
In the RLOO algorithm, the advantage function is closely tied to the reward signal. By collecting multiple online samples, the method enables unbiased variance reduction. Within this framework, each sample can serve as a baseline for the others. By subtracting this baseline from the reward of each sample, we compute its corresponding advantage value. In essence, this process evaluates how superior or inferior a given sample is relative to others in the batch, enabling more stable and efficient policy updates.

In our approach, we adopt a powered length penalty (PLP), where longer responses are penalized more mildly to allow for necessary reasoning on difficult questions. In contrast, methods like the proposed by \citet{arora2025training} use standardized length penalties, which normalize the penalty across samples. This normalization causes the reward values for longer responses to become nearly indistinguishable—especially when the length distribution is similar—thereby reducing the model’s ability to differentiate between long but meaningful outputs. Such methods tend to overly suppress long responses, even when they are essential for solving complex problems, which conflicts with our goal of encouraging deeper reasoning when needed. 

To characterize the statistical behavior of our length-based penalty, we consider a simplified setting where the sequence length \(\text{len}(y)\) is uniformly distributed over the interval \((a, b)\), i.e., \(\text{len}(y) \sim \mathcal{U}(a, b)\). In this study, we restrict our discussion to the case where $\gamma$ is equal to 0.5.
\[
Z = 1 + \frac{1}{\sqrt{\text{len}(y)}},
\]
with the scaling coefficient set to 1 for analytical convenience.
Under this assumption, the variance of \(Z\) is given by:
\begin{equation}
\text{Var}(Z) = \frac{\ln b - \ln a}{b - a} - \frac{4}{(\sqrt{a} + \sqrt{b})^2}.
\end{equation}

 Since the partial derivatives with respect to both $a$ and $b$ are negative, as both $a$ and $b$ increase, the variance decreases accordingly. That is, the goal is achieved of having a large difference in length penalties for simple problems and almost no difference in length penalties for difficult problems. However, the variance of the standardized samples remains consistently 1. Even with a penalty coefficient, the variance remains a fixed value and does not change with variations in length. The variances of the above two different methods also exhibit the same trend in general distributions.
 
We can visually observe from Figure \ref{fig:main} that when the length of the answer ranges from 300 to 600 and the answers are all correct, there are significant differences in the reward values of the two methods for answers of different lengths. However, when the length of the answer ranges from 7000 to 10000, our method imposes almost no length penalty on correct answers, while the efficient method still has a relatively large length penalty. Although the advantage differences among different answers are relatively small due to the small differences in reward values when our model gives long answers, this is the case when all answers are correct.
Once an error occurs in the answered question, that is, when a reward of 0 appears, the difference in the advantage value will emerge. This is in line with our goal when dealing with difficult questions, which is to place more emphasis on the correctness of the answer rather than the length of the answer.
\section{Experiments}
\begin{table*}[htbp]
    \centering
    \begin{tabular}{cccccccccc}
        \toprule
        \multicolumn{3}{c}{Models}& & \multicolumn{2}{c}{GSM8K} & \multicolumn{2}{c}{MATH500} & \multicolumn{2}{c}{AIME2024} \\
        \cmidrule(lr){1-4} \cmidrule(lr){5-6} \cmidrule(lr){7-8} \cmidrule(lr){9-10}
        \multicolumn{3}{c}{DeepSeek-R1} && Accuracy & Tokens & Accuracy & Tokens & Accuracy & Tokens \\
        \midrule
        \multirow{10}{*}{} & \multirow{5}{*}{1.5B} & Original & & 76.5\% & 720 & 82.9\% & 5446 & 27.7\% & 15643 \\
        % & & Efficient & & 85.4\% & 702 & 83.2\% & 3641 & 33.3\% & 15087 \\
        % & & Truncation & & 40.2\% & 371 &59.4\% &2312 &20.3\% &8676 \\
        & & PE  & &75.1\% &738 &83.1\% &5054 &33.3\% &15892  \\
        & & Efficient & & 85.4\% & 702 & 83.2\% & 3641 & 33.3\% & 15087 \\
        & & Ours & & \textbf{86.3\%} & \textbf{411} & \textbf{85.1\%} & \textbf{3606} & \textbf{33.7\%} & \textbf{13327} \\
        \cmidrule(lr){2-10}
        & \multirow{5}{*}{7B} & Original & & \textbf{92.6\%} & 1629 & \textbf{92.4\%} & 4141 & 54.7\% & 12816 \\
        % & & Efficient  & & 89.2\% & 226 & 90.7\% & 2329 & 48.7\% & 9721 \\
        % & & Truncation & &21.5\% &396 &67\% & 2329&44 \%&7975 \\
        & & PE & &87.7\% &619 & 91.4\%&3614 &49.7\% &13783 \\
        & & Efficient  & & 89.2\% & 226 & 90.7\% & 2329 & 48.7\% & 9721 \\
        & & Ours & & 90.1\% & \textbf{218} & 91.3\% & \textbf{1906} & \textbf{55.7\%} & \textbf{9056} \\
        \bottomrule
    \end{tabular}
        \caption{Model Performance Comparison, PE stands for Prompt Engineering. Efficient is the method used in \cite{arora2025training}. }
            \label{table:1}
\end{table*}
\subsection{Experiment Setup}
In this section,we provide the experiment results to evaluate the effectiveness of our method.\\
\textbf{Models} We conduct experiments on DeepSeek-R1 \cite{guo2025deepseek} and Qwen2.5 \cite{hui2024qwen2}. In our experiment we use three models from these families:DeepSeek-R1-Distill-Qwen-1.5B, Deepseek-R1-Distill-Qwen-7B and Qwen2.5-7B-Instruct. Among them, the first two are reasoning models and the Qwen2.5-7B-Instruct is non-reasoning model. We want to know whether our method can be effective for both reasoning and non-reasoning models.\\
\textbf{Datasets} The dataset used for training is GSM8K \cite{cobbe2021training}, it is a dataset of 8.5K high quality linguistically diverse grade school math word problems. The dataset was created to support the task of question answering on basic mathematical problems that require multi-step reasoning. For training, we selected 3,200 questions from the GSM8K training set. For each model, we generated 8 solutions for each question. The test data sets include GSM8K, MATH500 \cite{lightman2023let}, and AIME2024, covering a variety of mathematical problems at different difficulty levels.\\
\textbf{Baselines} In our experiment, we took the following methods as the baselines.\\
\text{(1) Prompt Engineering} In this approach, we incorporate prompts such as \textit{“Please output as little as possible.”} during the evaluation of the original model to shorten the length of the Chain-of-Thought.\\
\text{(2) Training Language Models to Reason Efficiently} This method standardizes the multiple answers for each sample, maps the resulting values to the range $[0,1]$ using the sigmoid function, and regulates the penalty strength via the coefficient $\alpha$.\\
\textbf{Implementation details} For the 1.5B model, we use 2 A100 GPUs for training, and for the 7B model, we use 8 A100 GPUs for training. We use vllm to limit the maximum length of the model's output during training. Since the dataset we use for training is the relatively simple GSM8K, we set the limit on the generation length to 2000 tokens. The training precision is set
to bfloat16. All models are trained with a batch size of 128 and for
every iteration we select 32 prompts from the dataset.Each prompt generates 8 responses. We set the learning rate for all models to  $5\times10^{-6}$. For each parameter setting, we perform three independent training runs and evaluate each model separately, reporting the average result. Each experimental run requires approximately two hours to complete.
\\
\textbf{Evaluation configurations} We follow the previous work and define the maximum generation length of all models as 37268 (including the thinking tokens and answer tokens). For Deepseek-like models, during the evaluation, we use model‘s official template. For each test question, we perform conditional sampling with a temperature of 0.6 and a top probability value of 0.95 to obtain N outputs, and then report the average accuracy of these $N$ outputs. Specifically, for GSM8K, which contains 1319 test samples, we set $N$ to 1, for MATH500, which has 500 samples, we set $N$ to 3, and for AIME2024, which only has 30 samples, we set $N$ to 10.
\subsection{Results}

% \begin{table*}[ht]
% \centering
%     \begin{tabular}{l|c|c} 
%       \textbf{Models} & \textbf{Accuracy} & \textbf{AVerage Tokens}\\
%       \hline
%         \multicolumn{3}{c}{GSM8K}\\
%       R1-Distill-Qwen-1.5B & 76.5\% & 720\\
%       R1-Distill-Qwen-1.5B(Efficient) & 85.4\% & 702\\
%       R1-Distill-Qwen-1.5B(ours) & 86.3\% & 411\\
%       R1-Distill-Qwen-7B & 92.6\% & 1629\\
%       R1-Distill-Qwen-7B(Efficient) & 89.5\% & 213\\
%       R1-Distill-Qwen-7B(ours) & 91.9\% & 139\\
%       R1-Distill-Qwen-7B(prompt enginering) & 87.7\% & 619\\
%       \hline
%         \multicolumn{3}{c}{MATH500}\\
%       R1-Distill-Qwen-1.5B & 82.9\% & 5446\\
%       R1-Distill-Qwen-1.5B(Efficient) & 83.2\% &3641 \\
%       R1-Distill-Qwen-1.5B(ours) & 85.1\% & 3606\\
%       R1-Distill-Qwen-7B & 92.4\% & 4141\\
%       R1-Distill-Qwen-7B(Efficient) & 90.7\% & 2329\\
%       R1-Distill-Qwen-7B(ours) & 91.3\% & 1906\\
%       \hline
%         \multicolumn{3}{c}{AIME2024}\\
%       R1-Distill-Qwen-1.5B & 27.7\% & 15634\\
%       R1-Distill-Qwen-1.5B(Efficient) & 33.3\% & 15087\\
%       R1-Distill-Qwen-1.5B(ours) & 33.7\% & 13327\\
%       R1-Distill-Qwen-7B & 54.7\% & 12816\\
%       R1-Distill-Qwen-7B(Efficient) & 52\% & 9003\\
%       R1-Distill-Qwen-7B(ours) & 55.7\% & 9056\\
        
%     \end{tabular}
% \caption{\label{citation-guide}
% Model efficiency results of Deepseek-like models.Among them, efficient refers to using the method of Training Language Models to Reason Efficiently under the same setting as ours on GSM8K.ours refers to the optimal model among the experimental parameters.
% }
% \end{table*}
\begin{table*}[ht]
\centering
    \begin{tabular}{l|c|c|c} 
            \toprule
      \textbf{Models} & \textbf{Accuracy} & \textbf{Average Tokens} &\textbf{Compression rate}\\
      \hline
        Original &91.4\% & 298 & 1.00 \\
        Tokenskip &89.9\% & 217& 0.70 \\
        Ours & \textbf{90.3\%} & \textbf{182}& \textbf{0.61} \\
        \bottomrule
    \end{tabular}
\caption{
Model efficiency results of Qwen2.5-7B-Instruct on GSM8K. The result of Tokenskip \cite{xia2025tokenskip}  refers to the situation where the compression ratio in the reference paper is 0.7.
}
\label{table:2}
\end{table*}

\begin{figure*}[htbp] % 使用 figure* 环境跨越两栏
    \centering
    \begin{subfigure}[b]{0.3\textwidth}
        \includegraphics[width=\textwidth]{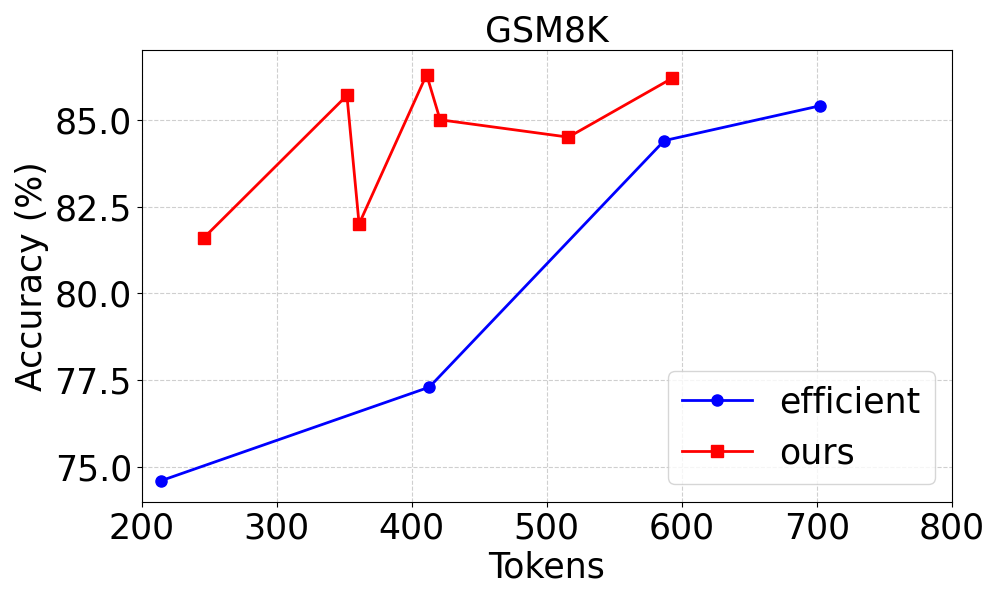}
        \caption{1.5B-GSM8K}
        \label{fig:sub1}
    \end{subfigure}
    \hfill  % 添加一些水平间距
    \begin{subfigure}[b]{0.3\textwidth}
        \includegraphics[width=\textwidth]{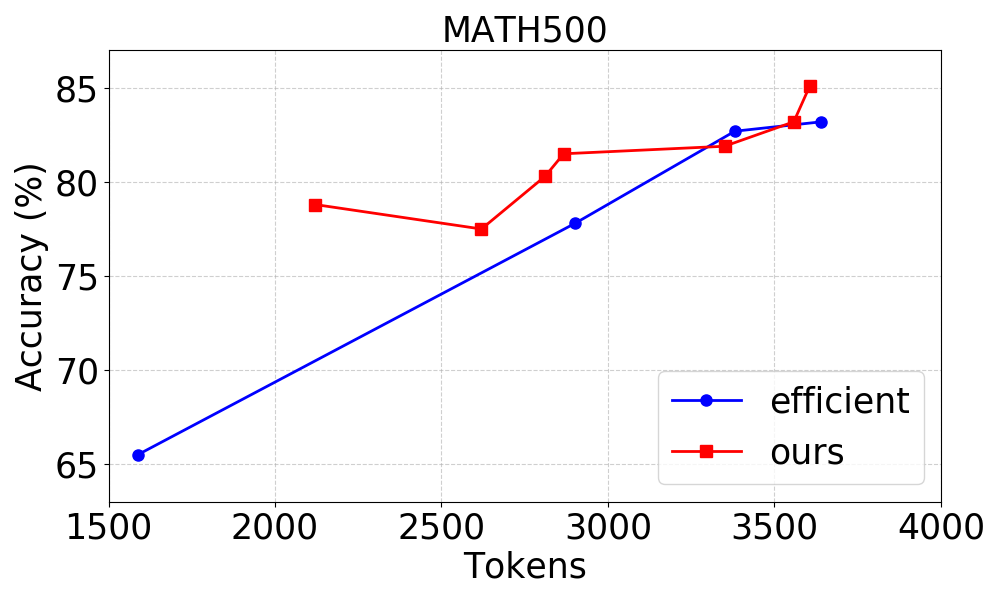}
        \caption{1.5B-MATH500}
        \label{fig:sub2}
    \end{subfigure}
    \hfill  % 添加一些水平间距
    \begin{subfigure}[b]{0.3\textwidth}
        \includegraphics[width=\textwidth]{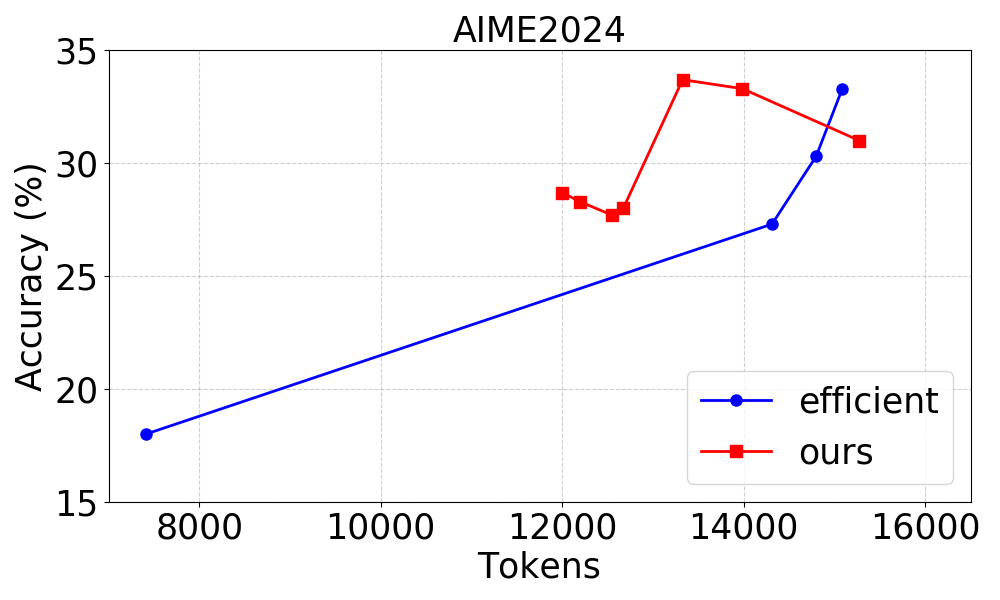}
        \caption{1.5B-AIME2024}
        \label{fig:sub3}
    \end{subfigure}
    \caption{Difference between our method and the efficient method. For our method, the coefficients are 1, 2, 3, 4, 5, 20, 30, while for the efficient method, the coefficients are 0.05, 0.1, 0.2, 0.4.
}
    \label{fig:5}
\end{figure*}

\begin{figure*}[htbp] % 使用 figure* 环境跨越两栏
    \centering
    \begin{subfigure}[b]{0.3\textwidth}
        \includegraphics[width=\textwidth]{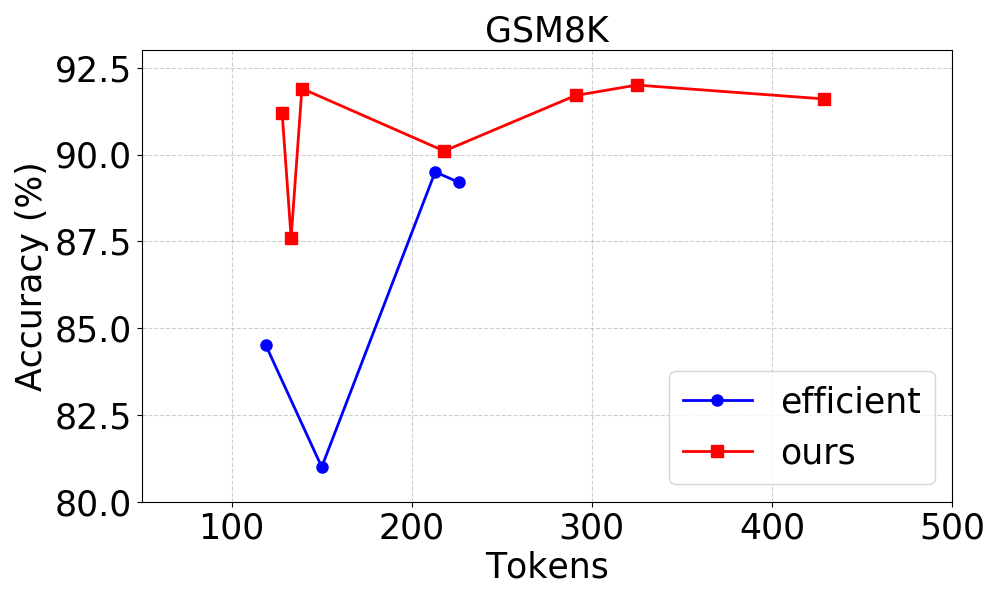}
        \caption{7B-GSM8K}
        \label{fig:sub1}
    \end{subfigure}
    \hfill  % 添加一些水平间距
    \begin{subfigure}[b]{0.3\textwidth}
        \includegraphics[width=\textwidth]{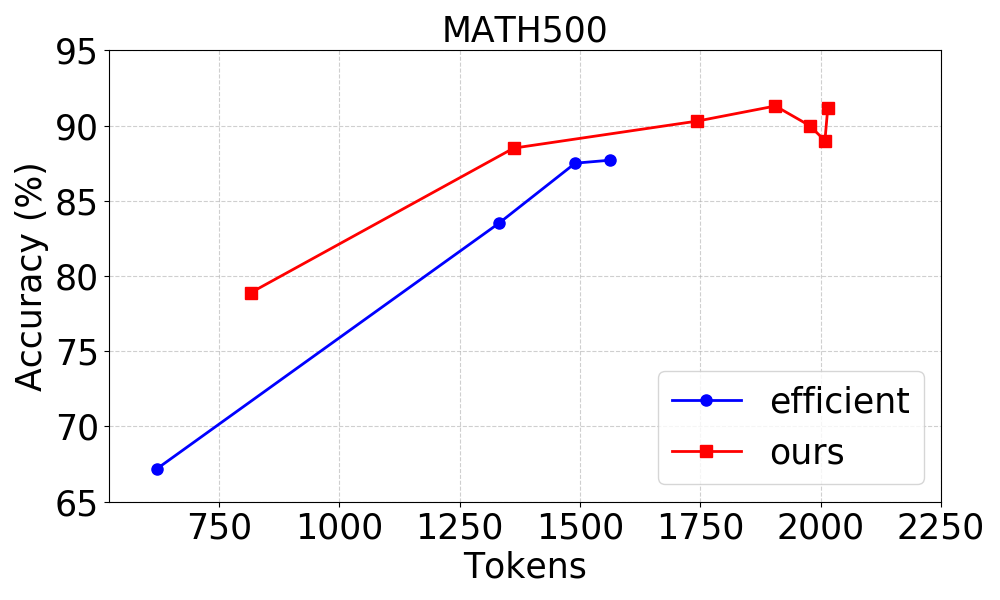}
        \caption{7B-MATH500}
        \label{fig:sub2}
    \end{subfigure}
    \hfill  % 添加一些水平间距
    \begin{subfigure}[b]{0.3\textwidth}
        \includegraphics[width=\textwidth]{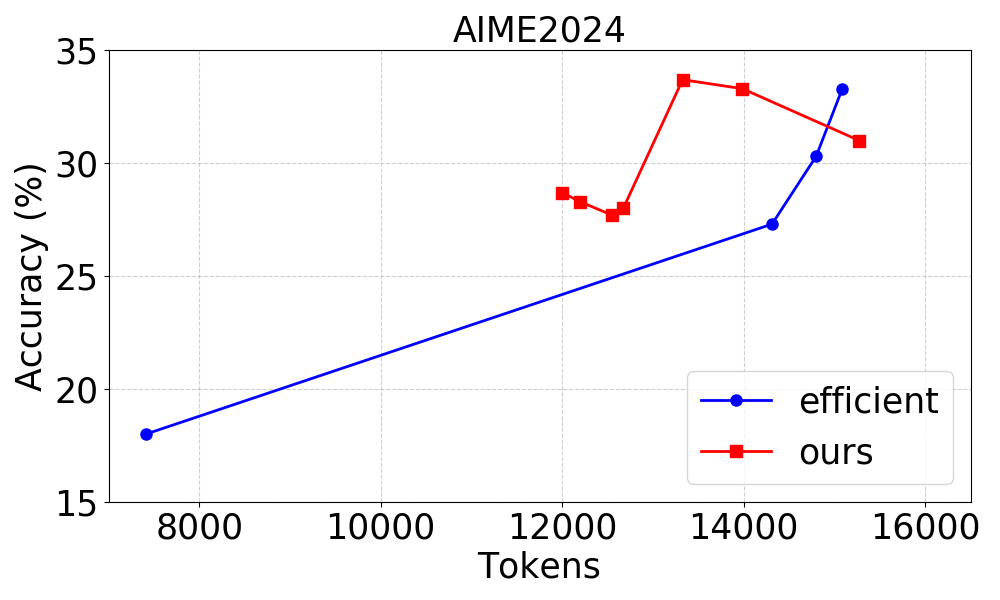}
        \caption{7B-AIME2024}
        \label{fig:sub3}
    \end{subfigure}
    \caption{Difference between our method and the efficient method. For our method, the coefficients are 1, 2, 3, 4, 5, 8, 10, while for the efficient method, the coefficients are 0.05, 0.1, 0.2, 0.4.
}
    \label{fig:6}
\end{figure*}

As shown in the Table \ref{table:1}, in the 1.5B model, the accuracy of our model in each dataset has increased compared to the original model. Moreover, the number of output tokens has also been reduced to a certain extent. For the 7B model, although the accuracy has decreased in the simple datasets, there is a remarkable reduction in the number of tokens. For instance, in GSM8K, the number of tokens has decreased from 1629 to 218, which is close to a 88\% reduction. And for the most difficult dataset AIME2024, the accuracy has slightly increased while the number of tokens has decreased.
Prompt engineering is often unstable, and in some tests, it even generates more tokens than not using it. 
correlated with the set range of forced truncation.
Although the efficient method can also improve the accuracy while reducing the number of tokens, its effect is not as good as ours.
And its stability is poor, as shown in the Figure \ref{fig:5} Figure \ref{fig:6}. For 1.5B model, when the value of a is set to 0.4, it can be observed that the number of tokens in the MATH500 and AIME2024 datasets drops sharply, accompanied by a significant decrease in accuracy. This situation also occurs in the 7B model. In comparison, for our model, when the number of tokens reaches the same order of magnitude, the decrease in accuracy is more gradual.

In Table \ref{table:2} we also conducted experimental comparisons on Qwen in GSM8K. For Qwen-like models, we used the Qwen2.5-7B-Instruct model to conduct experiments. We compared the original model, our method and the model trained by the Tokenskip method. It can be seen that for our model, while the accuracy only decreases by 1\%, the number of tokens is reduced by 40\%. Compared with TokenSkip, our method has a lower decrease in accuracy and a greater reduction in the number of tokens.
\begin{figure}[htbp] % 使用 figure* 环境跨越两栏
    \centering
    \begin{subfigure}[b]{0.4\textwidth}
        \includegraphics[width=\textwidth]{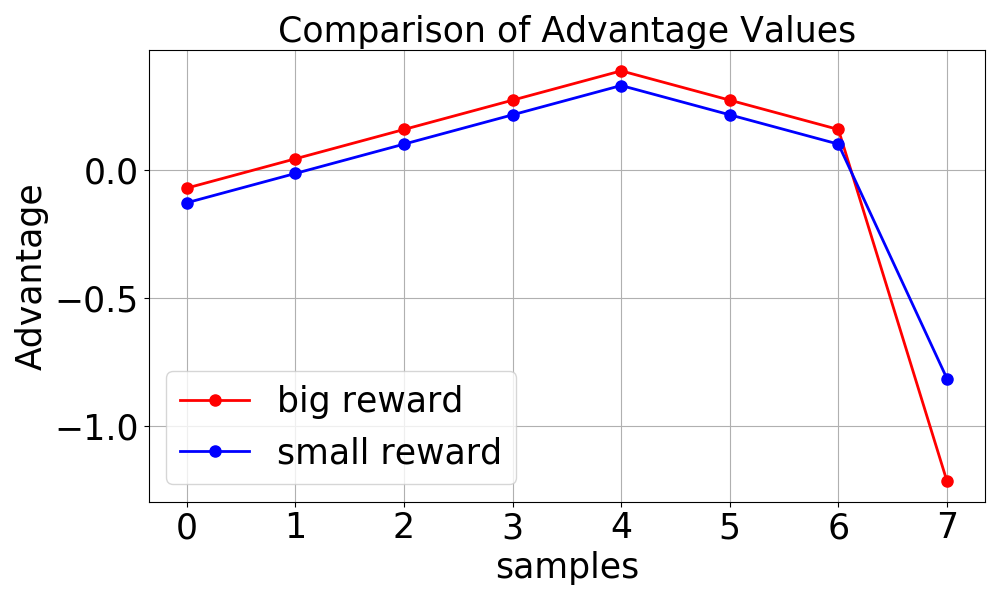}
    \end{subfigure}
    \caption{Difference between big reward and small reward when the last sample is incorrect.
}
    \label{fig:advantage3}
\end{figure}
\subsection{Empirical Analysis}
\textbf{Differences in model parameter scale} We have observed that, in both the 1.5B and 7B models, the reinforcement learning with length penalty reduced the number of tokens in the model's output. However, the extent of the reduction is not the same. It is obvious that in the 1.5B model, the length penalty brought by reinforcement learning is much smaller than that in the 7B model. We believe that this is because the 7B model is powerful enough, and the problems in the training dataset GSM8K are too simple for the 7B model. Therefore, even if the reasoning length is reduced by 90\%, it can still get the correct answer. This tendency has also led to a significant reduction in the reasoning chains for the MATH500 and AIME2024 datasets. This is why the optimization of the 7B model in terms of accuracy is not as significant as that of the 1.5B model.
\\
 \\
\noindent\textbf{Differences in length penalty strategies}
While most existing strategies impose a penalty by subtracting it from a constent such as 1 \cite{aggarwal2025l1,team2025kimi,xiao2025fast}, our approach introduces the penalty by adding it to the constant 1. Since the advantage function depends only on the relative differences among reward values within a group, rather than their absolute magnitudes, this modification achieves a comparable length penalty effect. We denote the two strategies as "big reward" and "small reward" for clarity. if we restrict both penalties to the intervals [0.6, 1] and [1, 1.4], the advantage values produced by the RLOO algorithm remain consistent when the underlying distributions coincide. However, in the event of an error—as exemplified by the eighth sample in Figure \ref{fig:advantage3}, where the reward falls to zero—our approach produces a substantially greater separation between correct and incorrect outputs than the efficient method. This larger separation explains why our accuracy remains relatively stable even as the token budget decreases markedly. By contrast, in the small-reward regime, an excessively high penalty coefficient can drive the reward of a long, correct answer toward zero. Consequently, some correct outputs incur disproportionately large penalties, diminishing their selection probability; once the answer length falls below the model’s capacity threshold, accuracy collapses sharply.\\
\noindent\textbf{Thought disappears} In this experiment, all the models of the Deepseek-R1 class we used are reasoning models. They will first engage in thinking before making an output. Then, in the form of <$\backslash$think><$\backslash$think>, they will use it as the content of their thinking before answering the questions. We are surprised to find that most responses in GSM8K do not have the label of the thinking process. However, for difficult questions, the model will still have the label of the thinking process.We statistically analyze the number of CoT in the original model and our model across three datasets of varying difficulty levels, as shown in Table \ref{table:3}. This means that the model has the ability to distinguish whether the question is simple enough to be answered correctly without going through the thinking process. This coincides with our expected goal. We compared the number of responses containing CoT between the original model and the trained model, and also statistically analyzed the proportion of CoT tokens to the total response tokens.
\begin{table}[htbp]
\centering
\begin{tabular}{l|c|c|c}
\toprule
\textbf{Models} & \textbf{GSM8K} & \textbf{MATH500} & \textbf{AIME2024} \\
\midrule
Original & 1319/1319 & 500/500 & 30/30 \\
Ours & \textbf{26}/1319  & \textbf{162}/500 & \textbf{30}/30 \\
\bottomrule
\end{tabular}
\caption{Comparison of our method with the original model on three mathematical benchmarks.}
\label{table:3}
\end{table}

% \begin{figure}[htbp] % 使用 figure* 环境跨越两栏
%     \centering
%     \begin{subfigure}[b]{0.5\textwidth}
%         \includegraphics[width=\textwidth]{emnlp2023-latex/cot.png}
%         \caption{}
%         \label{fig:sub3}
%     \end{subfigure}
%     \caption{This picture shows the proportional relationship between the length of thinking and the solution in responses.
% }
%     \label{fig:advantage4}
% \end{figure}
\subsection{Ablation Studies}

We conducted an ablation experiment. We carried out multiple experiments by selecting different parameters. As shown in Figure \ref{fig:5} and Figure \ref{fig:6},
On the 1.5B model, we selected parameters of 1, 2, 3, 4, 5, 20, and 30 for the experiment. As for the 7B model, we selected 1, 2, 3, 4, 5, 8, and 10 as parameters for the experiment. We found that as the value of $\alpha$ increases, the number of tokens generally shows a downward trend. In the 1.5B model, we found that selecting \(\alpha = 2\) yielded the best results, while in the 7B model, the optimal performance was achieved when \(\alpha = 4\). We have attempted to use other reinforcement learning algorithms, such as GRPO. However, the GRPO algorithm standardizes the rewards when calculating the advantage at the end. This results in small advantage values when all results are correct. After division by the standard deviation, the rewards are significantly magnified. Consequently, when the penalty term is added, the answer length decreases very rapidly during training. This causes the model to focus entirely on answer length rather than accuracy. Therefore, we decided to abandon this method.
\section{Conclusion}
In this paper, we proposed a method. By adding an absolute penalty for the answer length in the reward function, we can make the model's reasoning more efficient. That is, we can reduce or even eliminate the thought chain in simple questions, while paying more attention to the accuracy of the answers for difficult questions. We trained on the GSM8K dataset and tested on three datasets of different difficulty levels, namely GSM8K, MATH500, and AIME2024. The results show that our method shortens the answer length while maintaining almost the same accuracy on simple and medium datasets. For difficult datasets, although the reduction in answer length is relatively small, the accuracy improves compared with the original model.
\section{Limitations}
Due to resource constraints, the largest model we used is a 7B model, so our method has not been validated on models with larger parameters. Meanwhile, due to computational resource limitations, the generate number set during model training was 2000, which prevented us from using more difficult training datasets. In the future, we will use more resources to conduct experiments with the above settings. Moreover, we only tested on mathematical datasets. In the next step, we will attempt to test on datasets from other domains to observe whether the trained model has generalizability.

\bibliography{anthology}
\bibliographystyle{acl_natbib}
\cleardoublepage

\appendix
\section{Appendix}
\setcounter{table}{0}   %从0开始编号，显示出来表会A1开始编号
\setcounter{figure}{0}
\setcounter{section}{0}
\setcounter{equation}{0}
\section{Input Template}
In our training, we used different input templates for DeepseekR1-class models and Qwen-class models, as follows:\\
Input Template for DeepseekR1:\\
\textbf{<|begin\_of\_sentence|><|User|>Please reason step by step, and put your final answer within \textbackslash boxed\{\{\}\}.Question: \{\}<|Assistant|>}\\
\noindent{Input Template for Qwen:}\\
\textbf{<|im\_start|>system\textbackslash nYou are a helpful assistant.<|im\_end|>\textbackslash n<|im\_start|>user\textbackslash nPlease reason step by step, and put your final answer within \textbackslash\textbackslash boxed\{\{\}\}.\textbackslash n\{\}<|im\_end|>\textbackslash n<|im\_start|>assistant\textbackslash n}\\
\section{Results of Different Parameters }

We will present the results of all our hyperparameters in tabular form, where Models indicates the models used, Tokens represents the number of output tokens, and Acc denotes the accuracy. The same parameter was tested on GSM8K, MATH500, and AIME2024.
\begin{table*}[t]
    \centering
    % 第一行三张
    \begin{minipage}{0.3\textwidth}
        \centering
        \begin{tabular}{cccc}
        \toprule
        Model & $\alpha$ & Tokens & Acc \\
        \midrule
        R1-1.5B & 1 & 516 & 84.5\% \\
        R1-1.5B & 2 & 593 & 86.2\% \\
        R1-1.5B & 3 & 421 & 85.0\% \\
        R1-1.5B & 4 & 352 & 85.7\% \\
        R1-1.5B & 5 & 411 & 86.3\% \\
        R1-1.5B & 20 & 246 & 81.6\% \\
        R1-1.5B & 30 & 205 & 79.8\% \\
        \bottomrule
        \end{tabular}
        \captionof{table}{1.5B-GSM8K}
    \end{minipage}\hfill
    \begin{minipage}{0.3\textwidth}
        \centering
        \begin{tabular}{cccc}
        \toprule
        Model & $\alpha$ & Tokens & Acc \\
        \midrule
        R1-1.5B & 1 & 3558 & 83.2\% \\
        R1-1.5B & 2 & 3606 & 85.1\% \\
        R1-1.5B & 3 & 2867 & 81.5\% \\
        R1-1.5B & 4 & 2811 & 80.3\% \\
        R1-1.5B & 5 & 3350 & 81.9\% \\
        R1-1.5B & 20 & 2620 & 77.5\% \\
        R1-1.5B & 30 & 2569 & 74.6\% \\
        \bottomrule
        \end{tabular}
        \captionof{table}{1.5B-MATH500}
    \end{minipage}\hfill
    \begin{minipage}{0.3\textwidth}
        \centering
        \begin{tabular}{cccc}
        \toprule
        Model & $\alpha$ & Tokens & Acc \\
        \midrule
        R1-1.5B & 1 & 15275 & 31.0\% \\
        R1-1.5B & 2 & 13986 & 33.3\% \\
        R1-1.5B & 3 & 12000 & 28.7\% \\
        R1-1.5B & 4 & 12673 & 28.0\% \\
        R1-1.5B & 5 & 13327 & 33.7\% \\
        R1-1.5B & 20 & 12192 & 28.3\% \\
        R1-1.5B & 30 & 12391 & 21.7\% \\
        \bottomrule
        \end{tabular}
        \captionof{table}{1.5B-AIME2024}
    \end{minipage}

    \vspace{1em}

    % 第二行三张
    \begin{minipage}{0.3\textwidth}
        \centering
        \begin{tabular}{cccc}
        \toprule
        Model & $\alpha$ & Tokens & Acc \\
        \midrule
        R1-7B & 1 & 429 & 91.6\% \\
        R1-7B & 2 & 325 & 92.0\% \\
        R1-7B & 3 & 182 & 91.2\% \\
        R1-7B & 4 & 218 & 90.1\% \\
        R1-7B & 5 & 291 & 91.7\% \\
        R1-7B & 8 & 139 & 91.9\% \\
        R1-7B & 10 & 133 & 87.6\% \\
        \bottomrule
        \end{tabular}
        \captionof{table}{7B-GSM8K}
    \end{minipage}\hfill
    \begin{minipage}{0.3\textwidth}
        \centering
        \begin{tabular}{cccc}
        \toprule
        Model & $\alpha$ & Tokens & Acc \\
        \midrule
        R1-7B & 1 & 2015 & 91.2\% \\
        R1-7B & 2 & 1977 & 90.0\% \\
        R1-7B & 3 & 1744 & 90.3\% \\
        R1-7B & 4 & 1906 & 91.3\% \\
        R1-7B & 5 & 2009 & 89.0\% \\
        R1-7B & 8 & 1362 & 88.5\% \\
        R1-7B & 10 & 816 & 78.9\% \\
        \bottomrule
        \end{tabular}
        \captionof{table}{7B-MATH500}
    \end{minipage}\hfill
    \begin{minipage}{0.3\textwidth}
        \centering
        \begin{tabular}{cccc}
        \toprule
        Model & $\alpha$ & Tokens & Acc \\
        \midrule
        R1-7B & 1 & 9301 & 55.0\% \\
        R1-7B & 2 & 8383 & 53.7\% \\
        R1-7B & 3 & 10863 & 55.7\% \\
        R1-7B & 4 & 9058 & 55.7\% \\
        R1-7B & 5 & 8458 & 54.0\% \\
        R1-7B & 8 & 9618 & 50.7\% \\
        R1-7B & 10 & 7242 & 42.3\% \\
        \bottomrule
        \end{tabular}
        \captionof{table}{7B-AIME2024}
    \end{minipage}
    \label{fig:grouped_tables}
\end{table*}

\end{document}